\documentclass[a4paper]{article}
\pdfoutput=1
\usepackage{hyperref}
\usepackage{paralist}
\usepackage{enumitem}
\usepackage{INTERSPEECH2020}
\usepackage{graphicx}
\usepackage{subcaption}
\usepackage{lpic}
\usepackage{microtype}
\usepackage{cite}
\usepackage{booktabs}

\title{End-to-End Spoken Language Understanding Without Full Transcripts}
\name{Hong-Kwang J. Kuo, Zolt{\'a}n T\"{u}ske, Samuel Thomas, Yinghui Huang$^{*}$, Kartik Audhkhasi\sthanks{\ \ \ Work performed while at IBM}, \\ Brian Kingsbury, Gakuto Kurata, Zvi Kons, Ron Hoory, and Luis Lastras}
\address{IBM Research AI}
\email{}
\begin{document}

\maketitle
\begin{abstract}
An essential component of spoken language understanding (SLU) is slot filling: representing the meaning of a spoken utterance using semantic entity labels. In this paper, we develop end-to-end (E2E) spoken language understanding systems that directly convert speech input to semantic entities and investigate if these E2E SLU models can be trained solely on semantic entity annotations without word-for-word transcripts. Training such models is very useful as they can drastically reduce the cost of data collection. We created two types of such speech-to-entities models, a CTC model and an attention-based encoder-decoder model, by adapting models trained originally for speech recognition.   Given that our experiments involve speech input, these systems need to recognize both the entity label and words representing the entity value correctly. For our speech-to-entities experiments on the ATIS corpus, both the CTC and attention models showed impressive ability to skip non-entity words: there was little degradation when trained on just entities versus full transcripts. We also explored the scenario where the entities are in an order not necessarily related to spoken order in the utterance.  With its ability to do re-ordering, the attention model did remarkably well, achieving only about 2\% degradation in speech-to-bag-of-entities F1 score.
\end{abstract}
\noindent\textbf{Index Terms}: speech recognition, SLU

\section{Introduction}

Spoken language understanding is essential for a variety of applications including interactive spoken conversational systems and call center analytics that understands agent-customer dialogues.  
Slot filling is the process where we identify the entities (entity labels (e.g. fromloc.cityname) and values (e.g. Boston)).  This type of information is obviously important for completing transactions or information seeking requests.    

The ATIS (Air Travel Information Systems)~\cite{price1990evaluation,hemphill1990atis,tur2010left} corpus,  a publicly available corpus from the Linguistic Data Consortium, has been widely used for SLU research.
Initially, the best models for slot filling used Conditional Random Fields~\cite{raymond2007generative}, but more recently the best models use deep learning~\cite{liu2016attention,goo2018slot,liu2016joint,guo2014joint,kurata2016leveraging,liu2015recurrent,mesnil2014using,xu2013convolutional,hakkani2016multi}. Surprisingly, most ATIS studies used text transcripts as inputs. They were considered SLU simply because the text transcripts were from actual spoken utterances, and therefore were in a spoken style.  Only a few papers~\cite{mesnil2014using,liu2016joint,huangadapting} use the speech signal as input.  In this paper, we use speech inputs in an end-to-end spoken language understanding framework, taking speech as input and returning entity labels and values.

The goal of SLU is to understand the meaning of what was spoken, simplified in ATIS to an overall intent and a set of entities (slots).   In contrast with automatic speech recognition (ASR), where word for word accuracy is desired, SLU may not care about every word or even about how it was spoken (order of entities, word choices, etc.) as long as the meaning is preserved.   As a result, an SLU system may not need training data in the form of word-for-word transcripts, which are expensive to obtain for a new domain, assuming we are able to apply transfer learning using off-the-shelf general-domain ASR models previously trained on verbatim transcripts.

SLU systems have traditionally been a cascade of an automatic speech recognition (ASR) system converting speech into text followed by a natural language understanding (NLU) system that interprets the meaning of the text ~\cite{Goel2005,yaman2008integrative,Haghani2018}. In contrast, an end-to-end (E2E) SLU system  ~\cite{serdyuk2018towards,Haghani2018,qian2017exploring,chen2018spoken,ghannay2018end,lugosch2019speech,caubriere2019curriculum,huang2020leveraging} processes speech input directly into meaning without going through an intermediate text transcript. In this paper, we demonstrate that it is possible to train an end-to-end SLU system using a set (or bag) of entities that do not match the spoken order.  This may enable us to train on speech data from customer calls paired with transaction data produced by human agents. Imagine a human agent helps a client with a flight reservation, resulting in a transaction record containing the set of important entities.  This record could serve as light supervision for training the model we propose.  
By using just the speech recording and the bag of entities in training, we can drastically reduce the cost of data collection and thus increase the amount of training data.  Accurate verbatim transcription of speech data often requires 5-10$\times$ real-time for a human transcriber, not to mention additional costs for labeling entities.  In contrast, the transaction record containing the bag of entities is obtained during the course of helping the customer and has no additional cost.

\section{SLU use cases: what do entities look like?}
\label{sec:SLUentities}

For speech recognition, the training data is usually pairs of utterances and verbatim transcripts, as shown as (1) in the example below.  In order to train a slot filling model, such sentences need to be further labeled with entities, as shown in (2).  In this paper, we wish to train on speech that is paired with just the entities.  In (3), we use the entities presented in natural spoken order for training.  (3) differs from (2) simply in that all words that are not part of entities are excluded.  The entities can be thought of as the more important keywords; however, it does not mean that the other words do not carry any meaning.  For example, ``to'' and ``from'' clearly are important to determine whether a city is a destination or departure city.  In our model, such words will not be output, but the speech signal corresponding to those words will help the model to output the correct entity label.   Finally (4) makes the problem harder, but also more useful: the entities are not given in spoken order, but instead are sorted alphabetically according to entity name.  This simulates the semantic frame or bag of entities concept where the order of entities does not affect the meaning: \{\{fromloc.city\_name: RENO\}, \{stoploc.city\_name: LAS\_VEGAS \}, \{toloc.city\_name: DALLAS\}\}

\begin{enumerate}[leftmargin=*]
    \item {\bf Transcript:} i want a flight to dallas from reno that makes a stop in las vegas

    \item {\bf Transcript+entity labels:} {\it i want a flight to} DALLAS \mbox{\scriptsize B-toloc.city\_name} {\it from} RENO \mbox{\scriptsize B-fromloc.city\_name} {\it that makes a stop in} LAS \mbox{\scriptsize B-stoploc.city\_name} VEGAS \mbox{\scriptsize I-stoploc.city\_name}
    
    \item {\bf Entities in natural spoken order:} DALLAS \mbox{\scriptsize B-toloc.city\_name} RENO \mbox{\scriptsize B-fromloc.city\_name} LAS \mbox{\scriptsize B-stoploc.city\_name} VEGAS \mbox{\scriptsize I-stoploc.city\_name}
    
    \item {\bf Entities in alphabetic order:} RENO \mbox{\scriptsize B-fromloc.city\_name} LAS \mbox{\scriptsize B-stoploc.city\_name} VEGAS \mbox{\scriptsize I-stoploc.city\_name} DALLAS \mbox{\scriptsize B-toloc.city\_name}
        
\end{enumerate}

%
\section{Adapting ASR models into SLU systems}
\label{sec:training}
Given the different ways in which SLU data can be transcribed, we investigate various methods to train an SLU system. Starting from a pre-trained ASR model, we explore several design choices to understand how two different kinds of E2E models behave when used to model the various kinds of SLU data. Each possible training procedure employs one or more of the following steps.
\begin{enumerate}[leftmargin=*]
\item \textbf{ASR model adaptation to domain data (ASR-SLU adapt)}: Given that an off-the-shelf ASR model is likely trained on data that is acoustically different from the SLU data, a useful initial step is to adapt the ASR system. This step, which only uses verbatim transcripts, adapts the model to the novel acoustic conditions, words, and language constructs present in the SLU domain data. In model adaptation, one may use both the original general purpose ASR data (GP-ASR) and the domain data to provide better coverage of the ASR output units than adapting only on the domain data. 
\item \textbf{Joint ASR and SLU model training (joint ASR+SLU)}: In this step, entity labels are introduced into the training pipeline along with the full transcripts. This step is a form of curriculum learning~\cite{bengio2009curriculum,caubriere2019curriculum} that gradually modifies an off-the-shelf ASR model into a full fledged SLU model. What is novel in this step is that the model is now trained to output non-acoustic entity tokens in addition to the usual graphemic or phonetic output tokens.  For GP-ASR data, the targets are graphemic/phonetic tokens only, whereas for the SLU domain data, the targets also include entity labels.
Although this step is a natural progression in building the final SLU model, it can be skipped if sufficient SLU resources are available.
\item \textbf{SLU model fine tuning (fine tune SLU)}: In this final step, a model from step 1 or 2 above is fine tuned on just the SLU data to create the final SLU model. As described earlier, the entities that need to be recognized by the final SLU model might take different forms: within a full transcript, entities only in spoken order, or entities only in alphabetic order.
\end{enumerate}

\section{Building End-to-End SLU models}
\label{sec:exp}
Using the training procedure described above, we develop two variants of end-to-end SLU systems for the ATIS task that attempt to directly recognize entities in speech without intermediate text generation and text-based entity detection.

\subsection{SLU Data and Evaluation Metric}

We used the standard ATIS training and test sets for our experiments: 4978 training utterances from Class A (context independent) training data in the ATIS-2 and ATIS-3 corpora and 893 test utterances from ATIS-3 Nov93 and Dec94 data sets.  The entity labeled text data is found in LDC2019T04, but there are no pointers to corresponding audio files.  Only 518 (out of 893) test utterance audio files were found by~\cite{liu2016joint}.  We managed to find all 893 test audio files and 4976 (missing 2) training audio files, in the variety of spontaneous speaking mode, recorded with the Sennheiser microphone.

The 4976 training utterances comprise $\sim$9.64 hours of audio from 355 speakers. The 893 test utterances comprise $\sim$1.43 hours of audio from 55 speakers. The data was originally collected at 16~kHz, but we downsampled to 8~kHz to better model telephony use cases and so we could use off-the-shelf ASR models trained on conversational telephone speech. To better train the proposed E2E models, additional copies of the corpus are created using speed/tempo perturbation. The final training corpus after data augmentation is $\sim$140 hours of audio data. To simulate an additional  practical use case, we create a second \textit{noisy} ATIS corpus by adding street noise between 5-15~db SNR to the clean recordings. This $\sim$9.64 hours noisy  data set is also extended via data augmentation to $\sim$140 hours. A corresponding noisy test set is also prepared by corrupting the original clean test set with additive street noise at 5~db SNR.

We measure slot filling performance with the F1 score. When using speech input instead of text, word errors can arise. The F1 score requires that both the slot label and value must be correct.  For example, if the reference is {\em  toloc.city\_name:new\_york} but the decoded output is {\em  toloc.city\_name:york}, then we count both a false negative and a false positive. It is not sufficient that the correct slot label is produced: no ``partial credit'' is given for part of the entity value ({\em york}) being recognized. The scoring ignores the order of entities, and is therefore suitable for the ``bag-of-entities'' case we study. Our scoring script was tested on text-input systems and gave identical values as the standard scoring scripts.

\subsection{Evaluating CTC based SLU models}

To allow the SLU model to process entities and corresponding values independent of an external language model, we first construct a word CTC model on general purpose ASR data with the recipe steps presented in \cite{audhkhasi2019forget,kurata2019guiding} and using 300 hours of Switchboard (SWB-300) data. We then explore different training recipes to build CTC based SLU models.

Our first experiment assumes that we have both verbatim transcripts and entity labels for the SLU data and uses all three training steps. The \textbf{ASR-SLU adapt} step is performed as follows. The output layer of the ASR model, which estimates scores for 18,324 word targets and the blank symbol, is replaced with a randomly initialized output layer that estimates scores for 18,642 word/entity targets and the blank. The weights of the remaining 6 LSTM layers, each with 640 units per direction, and a fully connected bottleneck layer with 256 units are kept the same. The model is then trained on a combined data set of 300 hours of SWB GP-ASR data and 140 hours of clean ATIS data. Note that in this step, although the output layer has units for entity labels, the training targets are only words. In the \textbf{joint ASR+SLU} step, entity labels are introduced into the training transcripts and a joint ASR-SLU model is trained on the SWB+SLU data, starting from the final weights from the \textbf{ASR-SLU adapt} step. In the third and final \textbf{fine tune SLU} step, the joint ASR-SLU model is fine tuned on just the 140 hours of ATIS SLU data.

In experiment $[$1A$]$ of Table~\ref{tab:cleanATIS}, we evaluate this model on the clean test ATIS data. Given that the SLU model is a word CTC model, we do not use an external LM while decoding; instead, a simple greedy decode of the output is employed. This initial model has an F1 score of 91.7 for correctly detecting entity labels along with their values. In experiment $[$2A$]$, we develop a similar SLU model with full verbatim transcripts along with entity labels, but we skip the \textbf{ASR-SLU adapt} and \textbf{joint ASR+SLU adapt} steps. We initialize the model with the pre-trained SWB ASR model and directly train the SLU model. This model also achieves 91.7 F1 score, suggesting that the curriculum learning steps may not always be required.

\begin{table}[htpb]
    \caption{{\rm ATIS} bag-of-entities slot filling F1 score for speech input using CTC and Attention based models}
    \vspace{-2mm}
	\centering
	\begin{tabular}{@{}lccc@{}} \toprule
	\multicolumn{1}{c}{\bf Training Data} & \multicolumn{1}{c}{\bf Adapt} &  {\bf CTC} & \bf{Attention} \\ \cmidrule(lr){1-2}
    \cmidrule(lr){3-4}
		$[$1A$]$ Full transcripts & Y & 91.7 & 92.9 \\
		$[$2A$]$ Full transcripts & N & 91.7 & 93.0 \\ 
		$[$3A$]$ Entities, spoken order & Y & 92.7 & 92.8 \\
		$[$4A$]$ Entities, spoken order & N & 91.5 & 92.6 \\ 
		$[$5A$]$ Entities, alphabetic order & Y & 73.5 & 90.9 \\
		$[$6A$]$ Entities, alphabetic order & N & 61.9 & 90.6 \\
				\bottomrule
	\end{tabular}
	\label{tab:cleanATIS}
\end{table}

In the next set of experiments we investigate how important verbatim transcripts are for the training process. After the \textbf{joint ASR+SLU} step of experiment $[$1A$]$, in experiment $[$3A$]$, we train an SLU model that recognizes just the entity labels and their values in spoken order. We observe that the model learns to disregard words in the signal that are not entity values, while preserving just the entity values along with their labels. This model performs slightly better than full transcript model in $[$1A$]$. We extend this experiment in $[$4A$]$ by removing the use of transcripts entirely in the training process. This SLU model, after being initialized with a pre-trained ASR model, is trained directly to recognize entity labels and their values without any curriculum learning steps or verbatim transcripts. The model drops slightly in performance, but remains on par with the baseline systems. Finally, we train SLU systems on the much harder task of recognizing alphabetically sorted entity labels and their values. After the \textbf{joint ASR+SLU} step of experiment $[$1A$]$, in experiment $[$5A$]$ we train an SLU model that recognizes just the entity labels and their values, but now in alphabetic order. In experiment $[$6A$]$ a similar model is trained, but without any curriculum learning steps. On this task, the performance of the CTC model drops significantly as it is unable to learn from reordered targets. With the curriculum learning steps, the results in $[$5A$]$ are better, but still much worse than the baselines.

\subsection{Evaluating Attention based SLU models}
The attention models for SLU are initialized with a state-of-the-art ASR model developed for standard Switchboard ASR task. This model uses an encoder-decoder architecture in which the encoder is an 8-layer LSTM stack using batch-normalization, residual connections, and linear bottleneck layers~\cite{Hochreiter97,Ioffe2015,He2016,Vesely2011}.
The decoder models the sequence of BPE units estimated on characters~\cite{subword-nmt}, and consists of 2 unidirectional LSTM layers.
One is a dedicated language-model-like component that operates only on the embedded predicted symbol sequence, and the other jointly processes acoustic and symbol information.
The decoder applies additive, location aware attention~\cite{chorowski15}, and each layer has 768 unidirectional LSTM nodes.
As has been shown in~\cite{tuske20:arxiv}, exploiting various regularization techniques, including SpecAugment, sequence-noise injection, speed-tempo augmentation, and various dropout methods~\cite{Park2019,saon2019sequence,Ko15,hinton2012,pmlr-v28-wan13,Krueger2017}, results in state-of-the art speech recognition performance using this single-headed sequence-to-sequence model.

To recognize entities, the ASR model is adapted similar to the CTC model, following the steps of Section~\ref{sec:training}. In contrast to the CTC model, which uses word units, the attention model uses a smaller inventory of 600 BPE units and relies on the decoder LSTMs to model longer sequences --- the attention based model has an inherent long-span language model.
After the initial ASR model is trained on Switchboard, the subsequent adaptation and transfer learning steps used only the ATIS data without any Switchboard data.  Because the attention model operates at the sub-word level, and all new words appearing in the ATIS transcripts can be modeled using these sub-word units, no extension of the output and embedding layer is needed in the first \textbf{ASR-SLU adapt} step. 
We skip the \textbf{joint ASR+SLU} step and proceed directly to the \textbf{fine tune SLU} step, where the output and the embedding layers of the decoder must be extended with the entity labels. 
The softmax layer and embedding weights corresponding to the entity labels are randomly initialized, while all other parameters, including the weights which correspond to previously known symbols in the softmax and embedding layers, are copied over from the ASR model. Having no out-of-vocabulary words, sub-word level models are ideally suited to directly start the adaptation process with step 3 of Section~\ref{sec:training}.
All adaptation steps use 5 epochs of training.

The last column of Table~\ref{tab:cleanATIS} shows the slot filling F1 score for attention based SLU models. In experiment $[$1A$]$, an attention based ASR model trained on Switchboard-300h is first adapted on the clean ATIS data to create a domain specific ASR model.  On the test set, the word error rate (WER) using the base SWB-300 model is about 7.9\% which improved to 0.6\% after adaptation. This ASR model is then used as an initial model for transfer learning to create an SLU model. The F1 score is comparable to that of the CTC model.  In experiment $[$2A$]$, we skip the ASR adaptation step and directly use the SWB-300 ASR model to initialize the SLU model training.  In this scenario, there is no degradation in F1 score. There is no difference in SLU performance whether the model is  initialized  with a general purpose SWB-300 ASR model (WER=7.9\%) or with a domain adapted ASR model (WER=0.6\%).

We next consider the effects of training transcription quality or detail. Using transcripts that contain only entities in spoken order ($[$4A$]$), we obtain F1 scores that are almost the same as using full transcripts  ($[$1A$]$). When training transcripts contain entities in alphabetic order (possibly different from spoken order) ($[$6A$]$), there is a 2\% degradation in F1 score, from 92.9 to 90.9.  This result is much better than that for the CTC model (73.5), reflecting the re-ordering ability of attention based models.  As before, adding an extra step of ASR model adaptation ($[$3A$]$ and $[$5A$]$) with verbatim transcripts made little  difference.  This is encouraging, since we are assuming we are only given entities in the training transcripts.

Figure~\ref{fig:attn_alphabetic} shows the attention plots for the utterance {\em ``i would like  to  make  a reservation for a flight to denver from philadelphia on this  coming sunday''} with three different attention models:
\begin{inparaenum}[(a)]
\item ASR model,
\item SLU in spoken order, and 
\item SLU in alphabetic order.
\end{inparaenum}
The attention for (b) is largely monotonic with attention paid on consecutive parts of the audio signal corresponding to BPE units of keywords in the entities.  There are gaps reflecting skipping over non-entity words.  In (c), the attention is piece-wise monotonic, where the monotonic regions cover the BPE units within a keyword.  Since the entities are given in an order different from spoken order, the plot shows how the model is able to associate the correct parts of the speech signal with the entities.  In addition, at around 2 seconds, attention was paid to the phrase {\em ``make a reservation''} which is predictive of the overall intent of the sentence {\em ``flight.''}
(Intent recognition is left out of this paper for simplicity and due to lack of space.)

\begin{table}[t] %
	\caption{{\rm ATIS} bag-of-entities slot filling F1 score for speech input with additive street noise (5dB SNR)}
	\vspace{-2mm}
	\centering
	\begin{tabular}{@{}lccc@{}} \toprule
	\multicolumn{1}{c}{\bf Training Data} & \multicolumn{1}{c}{\bf Adapt} &  {\bf CTC} & \bf{Attention} \\ \cmidrule(lr){1-2}
    \cmidrule(lr){3-4}
		$[$1B$]$ Full transcripts & Y & 85.5 & 92.0 \\
		$[$2B$]$ Full transcripts & N & 79.6 & 91.3 \\ 
		$[$3B$]$ Entities, spoken order & Y & 88.6 & 91.2 \\
		$[$4B$]$ Entities, spoken order & N & 86.5 & 89.6 \\ 
		$[$5B$]$ Entities, alphabetic order & Y & 73.8 & 88.8 \\
		$[$6B$]$ Entities, alphabetic order & N & 68.5 & 87.7 \\
				\bottomrule
	\end{tabular}
	\label{tab:noisyATIS}
\end{table}

\begin{table}[t]%
	\caption{Effect of different amounts of data used to pre-train the ASR model used in initializing SLU model training}
	\vspace{-2mm}
	\centering
	\begin{tabular}{@{}lc@{}} 
	\toprule
	{\bf ASR Training} & \bf{Attention} \\ 
    \hline
		None & 78.1 \\
		Switchboard 300h & 92.6 \\ 
		Switchboard 2000h & 93.8 \\
	\bottomrule
	\end{tabular}
	\label{tab:dataAmt}
	\vspace{-2mm}
\end{table}

\subsection{Effect of Acoustic Mismatch}

In a next set of experiments (Table~\ref{tab:noisyATIS}), we use the \textit{noisy} ATIS corpus as the SLU data set and repeat the CTC based experiments conducted earlier. This set of experiments introduces additional variability to the training procedure with realistic noise in both training and test. Further, it increases the acoustic mismatch between the transferred model and the target domain. The general trends for the CTC model observed in Table~\ref{tab:cleanATIS} are also observed in Table~\ref{tab:noisyATIS}:
\begin{inparaenum}[(a)]
\item ASR transcript based curriculum training is effective; and,
\item entity labels can be recognized reasonably well in spoken order, but the performance is worse when the entity order is different.
\end{inparaenum}
In experiments like $[$2B$]$, the mismatch between the SLU data and the ASR data affects the performance of models that are only initialized with mismatched pre-trained models and have no other adaptation steps. The noise distortion in general causes these systems to drop in performance compared to the performance results in matched conditions.

Looking at the results in Table~\ref{tab:noisyATIS} for attention based SLU models in more detail, we note that there is an absolute degradation of 4.3\% in F1 score when we compare a model trained on full transcripts ($[$1B$]$ F1=92.0) to one trained on entities in alphabetic order ($[$6B$]$ F1=87.7\%).  While this is a significant drop in performance, it is much better than the CTC result of ($[$6B$]$ F1=68.5).  Compared to the clean speech condition, we also come to a different conclusion regarding the utility of ASR adaptation.  We see about 1\% improvement in F1 score when we are able to use an adapted ASR model instead of the base SWB-300 model to initialize SLU model training.  On the noisy test set, using the base SWB-300 model results in WER=60\%, whereas the ASR model adapted on noisy ATIS data gives WER=5\%.  It is remarkable that using these two very different ASR models to initialize the SLU model training leads to only a 1\% difference in F1 scores for the final models.

\subsection{Effect of the Amount of Pre-training Data}

Table~\ref{tab:dataAmt} shows how the amount of data used to train the ASR model for initializing SLU training affects the final F1 score.  Here we show only results for attention-based SLU models trained on entities in spoken order for clean speech.  We saw earlier that ASR adaptation on domain data does not always help.  But here, using 2000h instead of 300h for the initial ASR model improves the F1 score by about 1\%, 
most likely due to increased robustness of the model to unseen data: the unadapted WER on the ATIS test set is 3.1\% (SWB2000h) vs. 7.9\% (SWB300h). 
In contrast, when we directly train the SLU model from scratch, the best we could do was about F1=78.1.  When SLU data is limited, these experiments demonstrate the importance of ASR pre-training on a broad range of speech data, not necessarily related to the final SLU task.

\begin{figure}[t]
    \centering
\begin{lpic}{att00_asr(0.5)}
\lbl[tl]{0,50;{\footnotesize (a)}}
\end{lpic}
\begin{lpic}{att02_ordered(0.5)}
\lbl[tl]{0,38;{\footnotesize (b)}}
\end{lpic}
\begin{lpic}{att03_alphabetic(0.5)}
\lbl[tl]{0,39;{\footnotesize (c)}}
\end{lpic}

\caption{Attention plots for the utterance ``I would like  to  make  a reservation for a flight to Denver from Philadelphia on this  coming Sunday'': (a) ASR; (b) SLU in spoken order; (c) SLU in alphabetic order.
}
    \label{fig:attn_alphabetic}
\end{figure}

\section{Conclusions and Future Work}
\label{sec:conc}
In this paper we have investigated how various E2E SLU models can be built without verbatim transcripts. We have shown the importance of using pre-trained acoustic models and curriculum learning to build these systems. Using clean and noisy versions of ATIS, we explored the effects of entity order and acoustic mismatch on performance of these systems. This study shows that E2E systems can indeed be trained without verbatim transcripts and can predict entities reliably even if trained on transcripts where entities are not necessarily given in spoken order.
Our results provide useful insights to building better SLU systems in practical settings where full transcripts are often not available for training and the final SLU systems need to be deployed in noisy acoustic environments. The current study was limited to a setting with context independent utterances.  Future research may involve building SLU systems that operate on full conversations, rather than single utterances, where more complex linguistic phenomena like co-reference and entity linking are present.

\clearpage
\bibliographystyle{IEEEtran}
\bibliography{refs}
\end{document}